%% file: main.tex
\documentclass[10pt,twocolumn,letterpaper]{article}

\usepackage{iccv}              

\input{preamble}

\definecolor{iccvblue}{rgb}{0.21,0.49,0.74}
\usepackage[accsupp]{axessibility}
\usepackage[pagebackref,breaklinks,colorlinks,citecolor=iccvblue]{hyperref}
\usepackage{xcolor}
\usepackage[normalem]{ulem}
\usepackage{colortbl}
\usepackage{booktabs}
\usepackage{multirow}
\usepackage{tabularx}
\usepackage{graphicx}

\def\paperID{4934} 
\def\confName{ICCV}
\def\confYear{2025}

\title{BoxDreamer: Dreaming Box Corners for Generalizable Object Pose Estimation} 

\author{
Yuanhong Yu\textsuperscript{1,3} \quad
Xingyi He\textsuperscript{1,3} \quad
Chen Zhao\textsuperscript{4} \quad
Junhao Yu\textsuperscript{5} \quad
Jiaqi Yang\textsuperscript{6} \quad
Ruizhen Hu\textsuperscript{7} \\
Yujun Shen\textsuperscript{3} \quad
Xing Zhu\textsuperscript{3} \quad
Xiaowei Zhou\textsuperscript{1} \quad
Sida Peng\textsuperscript{1,2}\thanks{
    Corresponding author: Sida Peng. \\
}
\\
\textsuperscript{1}Zhejiang University\quad
\textsuperscript{2}Xiangjiang Laboratory\quad
\textsuperscript{3}Ant Group\quad \\
\textsuperscript{4}EPFL\quad
\textsuperscript{5}Chongqing University\quad 
\textsuperscript{6}Northwestern Polytechnical University\quad
\textsuperscript{7}Shenzhen University\quad
\\[0.5em] 
\href{https://zju3dv.github.io/boxdreamer}{https://zju3dv.github.io/boxdreamer}
}

\begin{document}
\maketitle
\input{sec/0_abstract}    
\input{sec/1_intro}

\input{sec/2_related_work}

\input{sec/3_method}
\input{sec/4_exps}

\input{sec/5_conclusion}

{
    \small
    \bibliographystyle{ieeenat_fullname}
    \bibliography{main}
}


\end{document}

%% file: preamble.tex
\definecolor{lightgreen}{RGB}{152, 251, 152}
\definecolor{lightpink}{RGB}{233, 23, 115}
\definecolor{purple}{rgb}{0.65,0,0.65}

\newcommand{\tocite}[1]{{\textcolor{red}{[TOCITE]}}}

\definecolor{best}{RGB}{255,179,179}        
\definecolor{secondbest}{RGB}{255,217,179}  
\definecolor{thirdbest}{RGB}{255,255,179}   
\newcommand{\bestcell}[1]{\cellcolor{red!30}\textbf{\strut #1}}


%% file: sec/0_abstract.tex
\begin{abstract}
This paper presents a generalizable RGB-based approach for object pose estimation, specifically designed to address challenges in sparse-view settings. While existing methods can estimate the poses of unseen objects, their generalization ability remains limited in scenarios involving occlusions and sparse reference views, restricting their real-world applicability. To overcome these limitations, we introduce corner points of the object bounding box as an intermediate representation of the object pose. The 3D object corners can be reliably recovered from sparse input views, while the 2D corner points in the target view are estimated through a novel reference-based point synthesizer, which works well even in scenarios involving occlusions. As object semantic points, object corners naturally establish 2D-3D correspondences for object pose estimation with a PnP algorithm. Extensive experiments on the YCB-Video and Occluded-LINEMOD datasets show that our approach outperforms state-of-the-art methods, highlighting the effectiveness of the proposed representation and significantly enhancing the generalization capabilities of object pose estimation, which is crucial for real-world applications.
\end{abstract}

%% file: sec/1_intro.tex
\section{Introduction}
\label{sec:intro}

\begin{figure}
    \centering
    \includegraphics[width=1.0\linewidth]{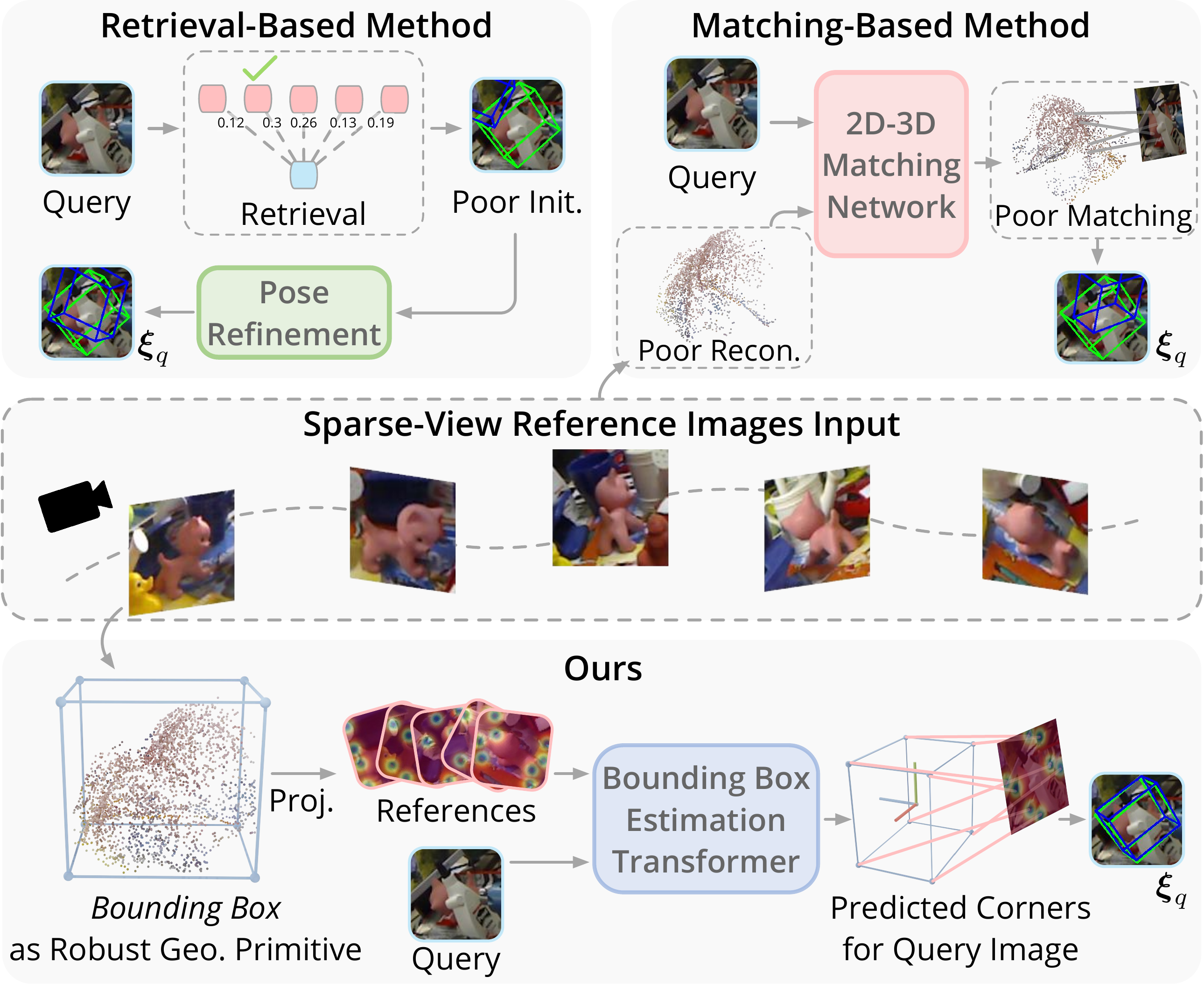}
    \caption{\textbf{Comparison of Different Generalizable Paradigms.} Unlike existing generalizable object pose estimation methods, BoxDreamer leverages a bounding box representation to handle incomplete object observations and achieves robust pose estimation even under severe occlusions.}
    \label{fig:teaser}
\end{figure}

Estimating the rigid transformation between an object and a camera, i.e., object pose estimation, is essential for diverse tasks such as augmented reality (AR) and object manipulation~\cite{chen2024g3flow, hsu2024spot}. 
We focus on generalizable object pose estimation from \emph{sparse-view} RGB images, where only a limited number of reference images are available for each object as prior information. 
This task requires the generalization ability of pose estimation methods to handle any unseen objects by a single forward pass and the robustness of the processes to accurately estimate poses from limited reference viewpoints. 
As shown in Fig.~\ref{fig:teaser}, current research on generalizable object pose estimation mainly includes two paradigms: \emph{Retrieval-based} approach~\cite{liu2022gen6d,zhao2024locposenet,cai_2024_GSPose,pan2024learning} and \emph{Matching-based} approach~\cite{sun2022onepose,he2022oneposeplusplus,castro2023posematcher}.

To estimate the target pose, \emph{Retrieval-based} methods initialize the object pose by retrieving the most similar reference image, followed by the post-refinement. These approaches are effective when dense reference views and corresponding pose annotations are available. 

However, for sparse-view scenarios, retrieving reference views with similar viewpoints is challenging, leading to inaccurate pose initialization.
The suboptimal pose initialization impairs subsequent modules that refine the pose.
Moreover, for objects in query images with severe occlusion, selecting the correct reference is highly challenging, often leading to the failure of the pose estimation process.

\emph{Matching-based} methods first reconstruct the target object's point cloud from reference images and then establish correspondences between the query image and the point cloud for pose estimation. These methods achieve high-accuracy pose estimation when accurate object point clouds can be recovered. However, the reliance on robust 2D-3D matching inherently depends on complete reconstructed point clouds, which becomes challenging in sparse-view scenarios. Additionally, object occlusion limits the association between image pixels and point clouds, significantly reducing the effectiveness of correspondence-based approaches in such conditions.

In this paper, we introduce a novel generalizable framework for 6D object pose estimation, which can predict the pose of any object using only a few reference images and can effectively handle severe occlusions.
As illustrated in Fig.~\ref{fig:teaser}, our key idea is to leverage the object bounding box as a geometric primitive, which can be efficiently reconstructed from sparse-view reference images. 
The 2D projections of the eight corners of the 3D bounding box are then accurately predicted in the query image for pose estimation. 
As semantic points of objects, the bounding box corners naturally establish 2D-3D correspondences, and both 3D and 2D corners can be recovered without dense views, making our method functions effective under sparse-view conditions.
In addition, compared with methods that require dense matching between the query image and the 3D primitive, our method is inherently more robust to occlusions.

Concretely, our approach involves two main steps: recovering the 3D bounding box and predicting 2D projections of the 3D bounding box corners for the query image.
First, we use sparse-view reconstruction tools to estimate camera poses and recover the object's approximate structure to compute its 3D bounding box.
To leverage 3D corner locations for object pose estimation, we project the 3D bounding box corners onto each reference image to generate corner heatmap, which are then fed into an end-to-end transformer decoder to predict the 2D projection of the 3D bounding box corners in the query view.
The global nature of object box corners enables our method to infer box corner positions based on the object's visible parts and reference demonstration. 
Finally, the 6DoF object pose is recovered using a Perspective-n-Point (PnP) algorithm with the 2D-3D correspondences.
Note that our method does not require an accurate 3D bounding box for object pose estimation, as demonstrated by our experimental results in Sec.~\ref{sec:exp-analysis}.

To evaluate the effectiveness of our proposed method, we conduct experiments on the Occluded LINEMOD~\cite{brachmann2014learning} and YCB-Video~\cite{xiang2018posecnn} datasets, which are rich in challenging scenarios like occlusion. Furthermore, we perform additional tests on diverse datasets, such as LINEMOD~\cite{hinterstoisser2012model} and the OnePose~\cite{sun2022onepose} dataset, to demonstrate our framework's generalization and sparse-view adaptability. Experimental results indicate that our method outperforms retrieval-based and matching-based methods under sparse-view conditions and shows more generalities and effectiveness under variable challenging scenarios.

In summary, this work has the following contributions:
\begin{itemize}[left=0pt, topsep=0pt, itemsep=0pt, parsep=0pt]
    \item A novel generalizable object pose estimation framework that effectively estimates object poses using only sparse RGB reference images.
    \item We propose object bounding box corners as an intermediate representation for generalizable object pose estimation.
    \item An end-to-end transformer decoder that directly regresses the 2D projection of the 3D bounding box in the query view with reference corner demonstration.
\end{itemize}

%% file: sec/2_related_work.tex
\section{Related Work}
\label{sec:related}

\subsection{CAD-Model-Based Object Pose Estimation}

Model-based object pose estimation methods can be broadly classified into two categories. \textbf{Instance-level pose estimators}~\cite{peng2019pvnet,xiang2018posecnn,park2019pix2pose,li2017deepim,rad2017bb8,dpod,epropnp,oberweger2018making,Li_2019_ICCV,kehl2017ssd, Hnig2024Improving2D} require separate training for each object and often rely on CAD models either to establish 2D--3xD correspondences~\cite{park2019pix2pose,peng2019pvnet,rad2017bb8,epropnp,oberweger2018making, Hnig2024Improving2D} or for rendering and pose refinement~\cite{li2017deepim,dpod}. There are also methods that attempt to regress object pose from RGB images directly.~\cite{xiang2018posecnn, Li_2019_ICCV,kehl2017ssd}. Although they achieve high accuracy for known objects, they are not generalizable to unseen objects. 

More generalizable, \textbf{category-level pose estimators}~\cite{wang2019normalized,ikeda2024diffusionnocs,wang2024gs,di2022gpv,you2022cppf, Zhang2024Omni6DL3, Cai2024OV9DOC, Chen2023SecondPoseSD, Zhang2023GenPoseGC, Lin2024CLIPoseCO, Zhang2024Omni6DPoseAB} first classify objects into predefined categories, then estimate their poses using category-specific priors. For example, NOCS~\cite{wang2019normalized} establishes a canonical coordinate system for each category from CAD models to provide category-level priors. Although larger datasets~\cite{Zhang2024Omni6DPoseAB, Zhang2024Omni6DL3} have enhanced generalizability, these methods still face challenges: in real-world scenarios, objects are too diverse or uncommon to be categorized, limiting their applicability for universal pose estimation.

Our framework addresses these challenges by removing the dependency on CAD models and instead leveraging the object's 3D bounding box. This approach is cost-effective and can be obtained from sparse-view, unposed RGB images, making it a robust solution for any-object pose estimation.

\subsection{CAD-Model-Free Object Pose Estimation}

To address the challenges of pose estimation for unseen objects, Gen6D~\cite{liu2022gen6d} introduced a framework leveraging a densely sampled view database with pose annotations for pose initialization and refinement, demonstrating strong generalizability without relying on CAD models. Subsequent methods~\cite{zhao2024locposenet,cai_2024_GSPose,pan2024learning} improved detection and refinement modules within this framework, but the dependency on densely captured views remains a significant limitation. Concurrently, OnePose~\cite{sun2022onepose} and its extension OnePose++~\cite{he2022oneposeplusplus} proposed reconstruction-based approaches that establish correspondences between reconstructed point clouds and query images. By incorporating Detector-Free SFM~\cite{he2024dfsfm}, OnePose++ further enhances accuracy, but these methods still require dense view captures, and the visibility and coverage of the captured images influence their performance. \


\subsection{Object 3D Representation}

Providing 3D information about the object is essential for generalizable object pose estimation. The OnePose family employs dense point clouds to represent objects, while Gen6D constructs a volumetric representation from multi-view images and utilizes it for pose refinement. LatentFusion~\cite{park2019latentfusion} establishes 3D latent to represent objects for pose estimation and refinement. GS-Pose~\cite{cai_2024_GSPose} represents objects using 3D Gaussians~\cite{kerbl20233d} and performs pose refinement based on it. Despite their advancements, none of these methods offer a simple yet robust 3D representation for 6-DoF pose estimation for any unseen object, as they heavily rely on dense reference images or struggle under challenging conditions such as occlusion or textureless surfaces. In contrast, our approach innovatively uses bounding box corners as a compact and effective representation for generalizable object pose estimation.

\subsection{Sparse-View Object Pose Estimation}
Sparse-view settings have attracted significant interest in the computer vision community due to their practicality and cost-efficiency~\cite{pan2024learning,wang2024gs, Nguyen_2024_CVPR,luo2024object,sun2024generalizable, Sun2024ExtremeTG, zhao20233d, zhao2024dvmnet}. In real-world applications, capturing dense and comprehensive object views is often time-consuming and resource-intensive. Therefore, accurately recovering object poses from sparse observations can greatly enhance real-time applications such as object manipulation.

Some recent works have addressed sparse-view pose estimation~\cite{luo2024object, Nguyen_2024_CVPR,sun2024generalizable}. For example, Luo et al.~\cite{luo2024object} employs an Object Gaussian representation with a render-and-refine strategy, achieving high accuracy but requiring about 10 minutes to construct the representation, which limits practical use. Meanwhile, Nguyen et al.~\cite{Nguyen_2024_CVPR} generates novel view embeddings, and Sun et al.~\cite{sun2024generalizable} leverages a novel view synthesis (NVS) model; however, both methods rely on generative performance, which may face challenges on real-world diverse objects, and NVS is also time-consuming. FoundationPose~\cite{wen2024foundationpose} shows impressive performance with sparse views through large-scale training on realistic synthetic datasets, yet its dependence on depth input restricts real-world use. In contrast, our method uses lightweight and effective representations for robust, real-time pose estimation from sparse-view references.

%% file: sec/3_method.tex
\section{Method}
\label{sec:method}

\begin{figure*}
    \centering
    \includegraphics[width=1.0\linewidth]{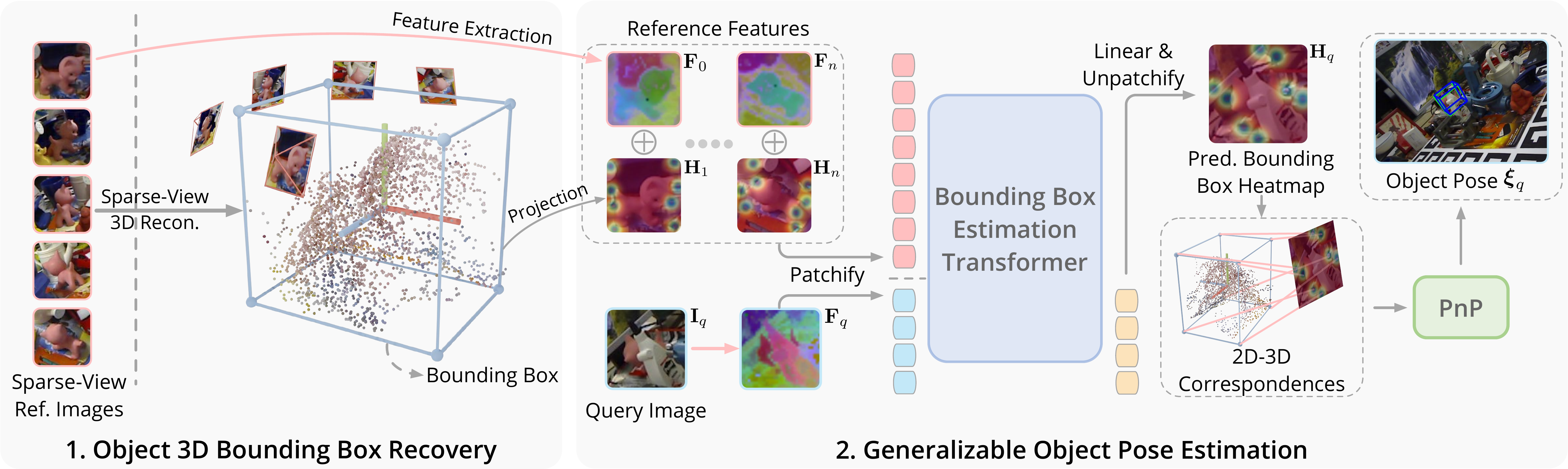}
    \caption{\textbf{Overview.} For each object, BoxDreamer first recovers its rough structure from a set of reference images using the sparse-view reconstruction method. During object pose inference, BoxDreamer predicts 2D bounding box heatmaps for the query image guided by reference box corners, establishing 2D-3D correspondences and recovering the object pose through the PnP algorithm.}
    \label{fig:pipeline}
\end{figure*}

The overview of BoxDreamer is illustrated in Fig.~\ref{fig:pipeline}.
Our method takes a query image $I_q$ and a set of reference images $\{I_0, \dots, I_i\}$ as input, along with the corresponding object detections $\{\mathcal{M}_0, \dots, \mathcal{M}_i\}$ to identify the object of interest, and aims to output the object pose $\mathbf{\xi}_q$.

To this end, we utilize an off-the-shelf reconstruction method to obtain the point cloud $\mathbf{P}$ of the target object, along with its 3D bounding box $\mathbf{B}$ and corresponding reference poses $\{\mathbf{\xi}_0, \dots, \mathbf{\xi}_i\}$ (see Sec.~\ref{sec:object_bounding_box_recognition}).

Then, we propose a novel bounding box estimation network to predict the projected box corners $\mathbf{b}_q$ in the query view. Based on these predictions, we first establish 2D-3D correspondences between the predicted 2D corners $\mathbf{b}_q$ and the reconstructed 3D corners $\mathbf{B}$, then recover the query object pose $\mathbf{\xi}_q$ using the PnP algorithm (see Sec.~\ref{sec:object_pose_estimation}).

\subsection{Preparation of Object Bounding Box}
\label{sec:object_bounding_box_recognition}
\noindent \textbf{3D Bounding Box as object representation.}
To address the limitations of OnePose, we avoid reliance on dense point clouds. Instead, we represent the object with a 3D bounding box $\mathbf{B}$, which serves as a rough spatial prior for pose estimation. The advantage of using a bounding box rather than a dense point cloud is that 3D bounding boxes can be restored from sparse view inputs, facilitating object coordinate recovery when dense observations are unavailable.

To obtain the 3D bounding box $\mathbf{B}$, we first collect sparse reference images. Using these images, we apply an off-the-shelf reconstruction method~\cite{dust3r_cvpr24, Leroy2024GroundingIM,wang2024vggsfm, Yang_2025_Fast3R} to extract the object's sparse geometry. We then filter out irrelevant points according to the object detection results, ultimately obtaining the 3D bounding box $\mathbf{B}$ that encloses the object.

By feeding cropped reference images to fully feed-forward reconstruction methods, we obtain the object point cloud $\mathbf{P}$ and the camera poses $\{\mathbf{\xi}_0, \dots, \mathbf{\xi}_i\}$ for the reference views. Specifically, for methods like DUSt3R~\cite{dust3r_cvpr24} (used in our experiments), sparse-view images are directly fed into the network to predict Pointmaps, which establishes dense correspondences between image pixels and associated 3D points. Once the Pointmap is obtained, the reference camera poses are efficiently recovered by applying the Perspective-n-Point (PnP) algorithm, leveraging the 2D-3D correspondences provided by the Pointmap. This process simultaneously recovers both 3D geometry and camera poses, making it particularly effective in sparse-view scenarios.

For each reference view with pose \(\mathbf{\xi} = (R,\mathbf{t})\), we first project a 3D point \(\mathbf{p} \in \mathbf{P}\) onto the image plane using the intrinsic matrix \(K\) with the perspective projection function \(\pi(\cdot)\):
\[
\mathbf{p}' = \pi\bigl(K (R\,\mathbf{p} + \mathbf{t})\bigr),
\]
where \(\pi(\cdot)\) converts homogeneous coordinates into 2D image coordinates.

Next, we discard points that fall outside the object region defined by the object detections $\{\mathcal{M}_0, \dots, \mathcal{M}_i\}$ in the reference images. The filtered point cloud is defined as:
\[
\tilde{\mathbf{P}} = \bigl\{\mathbf{p} \in \mathbf{P} \;\big|\; \forall i,\; \mathbf{p}' \in \mathcal{M}_i \bigr\}.
\]
Finally, the filtered point cloud \(\tilde{\mathbf{P}}\) is translated into an object-centric coordinate system and derives the 3D bounding box \(\mathbf{B}\) that encloses the object. 

\noindent \textbf{2D Heatmap as box representation.} We observed that the direct use of the eight corner coordinates of a 3D bounding box often degrades performance due to the inherent sparsity of the input signals. To better leverage the reference information, we project the 3D bounding box onto the image plane and generate a heatmap representing its 2D spatial distribution. This strategy aligns well with vision transformers, which can effectively process heatmap data.
Specifically, given the eight 3D corners of the bounding box \(\mathbf{B}\), we compute their projection onto the image plane $\mathbf{b}$ via the perspective projection function. 

Nevertheless, using a one-hot heatmap for these projected corners yields supervision signals that are excessively sparse and insufficiently smooth, which can hamper model learning. To mitigate this, we draw inspiration from CornerNet~\cite{Law_2018_ECCV}, which applies Gaussian smoothing around each ground-truth corner. This approach eases the penalty for negative locations within a certain radius and avoids sharp transitions between positive and negative regions. 

In practice, we observed that the hyperparameters from CornerNet were not directly transferable to our setting. To further enhance the smoothness of the heatmap, we redefine the heatmap function as:
\[
\mathbf{H}(x, y, i) = \exp\Bigl(-\frac{\sqrt{(x - x_i)^{2} + (y - y_i)^{2}}}{2\sigma^2}\Bigr),
\]
We set the denominator term $2\sigma^2$ to the square of one-tenth of the object size, which we define for each corner $i$ as its pixel distance to the object's 2D center.

By using 3D bounding boxes as spatial priors instead of dense point clouds, we design a novel pose estimation pipeline to better handle sparse-view inputs and incomplete reconstructions. This eliminates reliance on point cloud features and allows any suitable reconstruction approach to generate 3D bounding boxes, yielding a more flexible pipeline. Furthermore, representing bounding boxes via a 2D heatmap avoids direct dependence on potentially noisy or scale-sensitive 3D points.

\subsection{Object Pose Estimation}
\label{sec:object_pose_estimation}
After obtaining the bounding box heatmaps for the reference views $\{\mathbf{H}_0, \dots, \mathbf{H}_i\}$, we employ an end-to-end transformer decoder to directly infer the corresponding heatmap $\mathbf{H}_q$ for the query view.

More precisely, we feed both the reference and query images into a pre-trained DINOv2~\cite{oquab2023dinov2} model to extract image features
$\{\mathbf{F}_0, \dots, \mathbf{F}_i, \mathbf{F}_q\} \in \mathbb{R}^{\frac{H}{p} \times \frac{W}{p} \times d}$, where $p$ is the patch size and $d$ is the feature dimension.

Next, we divide each bounding box heatmap $\mathbf{H}_i \in \mathbb{R}^{H \times W \times 8}$ into non-overlapping patches, yielding a patched heatmap $\mathbf{H}_i^p \in \mathbb{R}^{\frac{H}{p} \times \frac{W}{p} \times 8p^{2}}$. 
To fuse the image features with the patched heatmap, we apply a linear layer to project the heatmap tokens to the same dimension as the image features:
\[
\mathbf{H}^p_i = \text{Linear}\bigl(\mathbf{H}^p_i\bigr) \in \mathbb{R}^{\frac{H}{p} \times \frac{W}{p} \times d}.
\]
We then add the projected heatmap tokens element-wise to the corresponding image features:
\[
\mathbf{F}_i' = \mathbf{F}_i + \mathbf{H}^p_{i}.
\]
For the query image, rather than heatmap tokens, we use learned query tokens $\mathbf{Q} \in \mathbb{R}^{\frac{H}{p} \times \frac{W}{p} \times d}$. We then flatten the reference and query features and concatenate them along the patch dimension, producing a 1D token sequence with length $l = (N + 1) \times \frac{H \times W}{p^2}$, where $N$ is the number of reference views. 
This sequence is then fed into a full self-attention transformer decoder comprising $L$ layers, generating the query bounding box features $\mathbf{F}_q' \in \mathbb{R}^{\frac{H}{p} \times \frac{W}{p} \times d}$.

Finally, we apply a linear layer to map the query features back to the original heatmap dimension and unpatchify the result to obtain the final query bounding box heatmap:
\[
\mathbf{H}_q = \text{Sigmoid}\Bigl(\text{Linear}(\mathbf{F}_q')\Bigr) \in \mathbb{R}^{H \times W \times 8}.
\]

Given the predicted query bounding box heatmap and the reconstructed 3D bounding box, we establish 2D-3D correspondences by assigning each heatmap channel to a specific bounding box corner according to a predefined channel-wise order. We then recover the object pose $\mathbf{\xi}_q$ using the Perspective-n-Point (PnP) algorithm.

\subsection{Training}

\textbf{Supervision.} We use Smooth L1 Loss~\cite{girshickICCV15fastrcnn} to supervise both the predicted bounding box heatmap (\textit{coarse}) and each specific corner point (\textit{fine}). 

The coarse loss is defined as:
\[
L_{\text{coarse}} 
= \frac{1}{N} \sum_{i=1}^{N} \text{SmoothL1}\bigl(h_i,\, \hat{h}_i\bigr),
\]
where \(h_i\) and \(\hat{h}_i\) denote the ground-truth and predicted heatmap values, respectively.

The fine loss is designed to improve the accuracy of corner points:
\[
L_{\text{fine}} 
= \frac{1}{8} \sum_{i=1}^{8} \text{SmoothL1}\bigl(b_i,\, \hat{b}_i\bigr),
\]
where $b_i$ and $\hat{b}_i$ are the ground-truth and predicted corner coordinates, respectively. 
The final loss combines the coarse and fine losses:
\[
L = L_{\text{coarse}} + \lambda\, L_{\text{fine}},
\]
Where $\lambda$ is a hyperparameter balancing the two terms, we set $\lambda = 2.0$ in our experiments.

\subsection{Implementation Details} 
\label{sec:implementation_details}

\noindent \textbf{Training Data.} We use the Objaverse (synthetic)~\cite{objaverse,objaverseXL} and OnePose (real)~\cite{sun2022onepose} datasets for training, which consist of over 45k synthetic objects, 50 real objects, and a total of 2.9M+ images.

\noindent \textbf{Data Augmentation.} To boost the generalizability, we randomly rotate the 3D bounding box (within $[-\pi, \pi]$ along a random axis) to break its direct association with semantic information. We also apply RGB augmentations such as motion blur and noise, and composite synthetic images with random SUN2012~\cite{xiao2010sun} backgrounds. Moreover, we further improve occlusion handling by randomly occluding objects through truncation and masking.

\noindent \textbf{Network Architecture.} The network leverages a transformer decoder consisting of $L=12$ layers, $d=768$ hidden size, and $8$ attention heads. The patch size is set to $14$ following the DINOv2-base~\cite{oquab2023dinov2} model. 

\noindent \textbf{Training.}
The model is trained using the AdamW optimizer with an initial learning rate of $10^{-4}$, following a cosine decay schedule. During training, the number of reference images is dynamically sampled between $1$ and $15$, while the batch size per GPU varies from $144$ to $18$. The model is trained for 100 epochs using 8 A100-SXM4-80GB GPUs.

%% file: sec/4_exps.tex
\section{Experiments}
\begin{figure*}
    \centering
    \includegraphics[width=\textwidth]{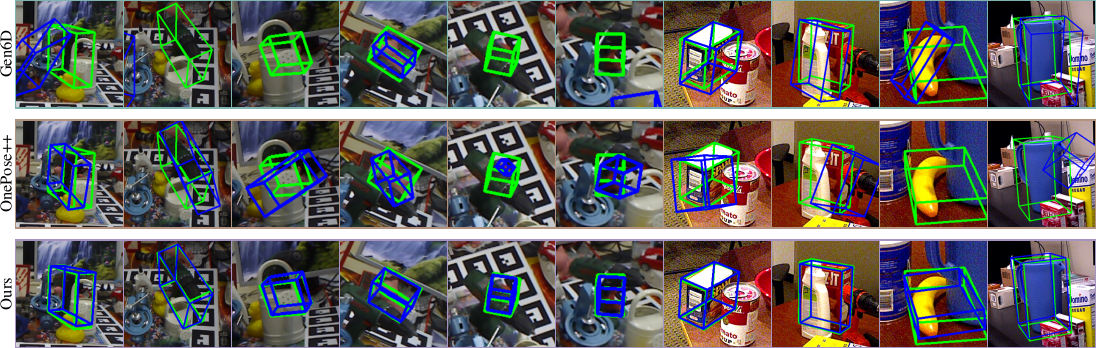}
    \caption{\textbf{Qualitative Comparison on Occluded-LINEMOD and YCB-Video.} Green boxes indicate the ground truth, while blue boxes represent the predicted results. Both quantitative and qualitative results demonstrate the method's effectiveness in occlusion.}
    \label{fig:qualitative}
\end{figure*}

\label{sec:exps}

In this section, we mainly evaluate our method across four datasets: LINEMOD and Occluded LINEMOD (Sec.~\ref{sec:linemod}), YCB-Video (Sec.~\ref{sec:ycb}) and OnePose-LowTexture (Sec.~\ref{sec:onepose}). 
We also evaluate our method's generalizability on OnePose, and correspondent results can be found in the supplementary material.
We begin by introducing the experimental setup and baseline selection in Sec.~\ref{sec:setup}.

\subsection{Experimental Setup and Baselines}
\label{sec:setup}

\textbf{Baselines.} We evaluate our method against two representative paradigms for generalizable pose estimation: Gen6D and OnePose++. While LocPoseNet~\cite{zhao2024locposenet} improves the detection accuracy of Gen6D, we use accurate ground-truth detection results for Gen6D in all our experiments to ensure precise initial translation. Therefore, a direct comparison with LocPoseNet is unnecessary. 

In addition, for recent methods such as Cas6D~\cite{pan2024learning} and GS-Pose, which also provide sparse-view results and use the same view sample strategy as our method, we include a comparison with them in Sec.~\ref{sec:linemod}.

To further demonstrate the effectiveness of our approach, we also compare it against state-of-the-art instance-level pose estimation methods~\cite{peng2019pvnet,xiang2018posecnn,rad2017bb8}.

\noindent \textbf{Reference Databases.}  
We utilize the following reference databases for evaluation:
\begin{itemize}
    \item \textbf{LINEMOD and Occluded LINEMOD}: We use the standard training-testing split from prior studies~\cite{peng2019pvnet,liu2022gen6d,sun2022onepose,he2022oneposeplusplus}, with $\sim$180 reference images per object. 
    \item \textbf{YCB-Video}: We create three distinct reference databases to assess robustness across varying reference qualities:
    (a) Sparse Database: Contains 16 reference images per object, sampled from distinct video sequences as in FoundationPose~\cite{wen2024foundationpose}.
    (b) Most-overlapping Database: To best test baseline methods, we build this database by selecting the most overlapping video sequence as the reference database.
    (c) Occlusion-minimizing Database: To mitigate occlusion in the most overlapping sequence, we manually select a reference sequence that maintains high overlap with minimal occlusion.
    Details on the construction of the YCB-Video reference databases are provided in the supplementary material.  
    \item \textbf{OnePose and OnePose-Lowtexture}: We use the first sequence as the reference database, as in~\cite{sun2022onepose, he2022oneposeplusplus}.
\end{itemize}

To ensure consistency across all datasets and methods, we uniformly sample sparse-view references using the Farthest Point Sampling (FPS) algorithm. For experiments with more than five reference images, unless stated otherwise, BoxDreamer initially selects five neighbors by directly computing the cosine similarity between the query and the reference DINOv2 features.

\noindent \textbf{Metrics.} The evaluation metrics are selected following established practices in prior works for each dataset:
\begin{itemize}
    \item ADD(s)-0.1d and Proj2D (with a 5-pixel threshold) are used for LINEMOD, Occluded LINEMOD, and the subset of OnePose-LowTexture with available CAD models. 
    \item ADD-AUC and ADDs-AUC serve as the metrics for the YCB-Video dataset.  
\end{itemize}

\subsection{Results on LM and Occluded LM}
\label{sec:linemod}

\begin{table}
    \centering
    \resizebox{0.7\columnwidth}{!}{
    \begin{tabular}{ c c c c c }
        \hline
        \textbf{Method} & \multicolumn{4}{c}{\textbf{Number of References} ($N_r$)} \\ \cline{2-5}
             & 5 & 16 & 25 & 32 \\ \hline
             OnePose++ & 1.1 & 31.4 & 48.6 &\cellcolor{thirdbest}55.0 \\
             Gen6D$^{\dagger}$ &\cellcolor{thirdbest}18.0& - &\cellcolor{thirdbest}58.6& - \\
             Gen6D$^{\ddagger}$ & - & 29.1 & - & 49.4 \\
             Cas6D & - &\cellcolor{thirdbest}32.4& - & 53.9 \\
             GS-Pose$_{init}$ & 4.5 & 15.6 & 21.1 & 23.4 \\
             GS-Pose &\cellcolor{secondbest}25.6&\bestcell{62.1}&\bestcell{69.4}&\bestcell{74.5} \\
             \hline
                    \textbf{Ours} &\bestcell{53.1}&\cellcolor{secondbest}60.4&\cellcolor{secondbest}65.9&\cellcolor{secondbest}69.2 \\
             \hline
    \end{tabular}
    }
    \caption{
        \textbf{Comparison on the LINEMOD subset.} ADD(s)-0.1d is reported. Gen6D$^{\dagger}$ indicates that all reference images are used for refinement, and Gen6D$^{\ddagger}$ is the public results from GS-Pose, GS-Pose$_{init}$ reports the initial results without Gaussian Splatting refinement. We've highlighted the \colorbox{best}{\textbf{best}}, \colorbox{secondbest}{second-best}, and \colorbox{thirdbest}{third-best} performing metrics in the table.
    }
    \label{tab:LINEMOD-Avg}
\end{table}

We first evaluate the proposed method on two widely used object pose estimation benchmarks: LINEMOD and Occluded LINEMOD. Experiments are conducted under two distinct levels of reference database density, namely, five reference views and 25 reference views. In the latter case, the number of reference views is determined as the minimum requirement for OnePose++ to reconstruct all objects successfully. For LINEMOD, additional experiments are performed using databases containing 16 and 32 reference views to compare with the results provided by GS-Pose and Cas6D. Detailed comparisons with Gen6D and OnePose++ can be found in the Supplementary Material.

Table~\ref{tab:LINEMOD-Avg} summarizes performance on the LINEMOD subset. Note that Gen6D was trained on a different subset, and for Cas6D, we use the results reported in its paper since no public checkpoints are available. For GS-Pose, we retested performance using the same reference databases as ours, with refinement parameters set to the default values from the public source code.

With only five reference images, our method outperforms all competitors, achieving twice the performance of the second-best GS-Pose. As more reference images are used, our approach matches GS-Pose's overall performance while surpassing its initial results. Notably, our method is over 40 times faster than GS-Pose's 0.96s per query image during refinement.
\begin{table}[htbp]
    \centering
    \resizebox{\columnwidth}{!}{
    \begin{tabular}{c c  cccccccc  c}
    \toprule
    \multirow{2}{*}{\textbf{Ref. images}} & \multirow{2}{*}{\textbf{Method}} & \multicolumn{8}{c}{\textbf{Objects}} & \multirow{2}{*}{\textbf{Avg.}} \\ \cline{3-10}
    & & \textbf{ape} & \textbf{can} & \textbf{\textit{cat}} & \textbf{\textit{driller}} & \textbf{\textit{duck}} & \textbf{eggbox$ ^{*}$} & \textbf{glue$ ^{*}$} & \textbf{holepuncher} &  \\ \midrule
    & & \multicolumn{8}{c}{\textit{ADD(s)-0.1d}} & \\ \midrule
    \multirow{2}{*}{-}
    & PoseCNN & 9.6 & 45.2 & 0.9 & 41.4 & 19.6 & 22.0 & 38.5 & 22.1 & 24.9 \\
    & PVNet  & 15.8 & 63.3 & 16.7 & 25.2 & 65.7 & 50.2 & 49.6 & 39.7 & 40.8 \\ 
    \midrule
    \multirow{4}{*}{5}
    & OnePose++ & - & 0.0 & 0.0 & 0.0 & - & - & - & 0.0 & - \\
    & Gen6D &\cellcolor{thirdbest}5.2&\cellcolor{thirdbest}8.6&\cellcolor{thirdbest}0.3&\cellcolor{thirdbest}0.3&\cellcolor{thirdbest}5.2&\cellcolor{thirdbest}5.8&\cellcolor{thirdbest}7.4&\cellcolor{thirdbest}4.8&\cellcolor{thirdbest}4.7 \\
    & Gen6D$^{\dagger}$ &\cellcolor{secondbest}6.4&\cellcolor{secondbest}8.8&\cellcolor{secondbest}0.5&\cellcolor{secondbest}4.8&\cellcolor{secondbest}5.4&\cellcolor{secondbest}29.6&\cellcolor{secondbest}15.0&\bestcell{11.1}&\cellcolor{secondbest}10.2 \\
    & \textbf{Ours} &\bestcell{21.5}&\bestcell{48.4}&\bestcell{4.0}&\bestcell{56.9}&\bestcell{16.1}&\bestcell{34.7}&\bestcell{20.6}&\cellcolor{secondbest}9.2&\bestcell{26.5} \\
    \midrule
    \multirow{4}{*}{25}
    & OnePose++ & 0.0 & 0.2 & 2.6 & 0.0 & 1.0 & 21.5 & 2.6 & 2.2 & 3.8 \\
    & Gen6D &\cellcolor{thirdbest}12.3&\cellcolor{thirdbest}23.4&\cellcolor{thirdbest}5.6&\cellcolor{thirdbest}2.5&\cellcolor{secondbest}17.2&\cellcolor{thirdbest}25.2&\cellcolor{thirdbest}13.4&\cellcolor{thirdbest}28.5&\cellcolor{thirdbest}16.0 \\
    & Gen6D$^{\dagger}$ &\cellcolor{secondbest}14.2&\cellcolor{secondbest}29.9&\cellcolor{secondbest}7.4&\cellcolor{secondbest}21.1&\cellcolor{thirdbest}15.4&\bestcell{45.8}&\cellcolor{secondbest}31.1&\cellcolor{secondbest}38.9&\cellcolor{secondbest}25.5 \\
    & \textbf{Ours}  &\bestcell{21.7}&\bestcell{61.6}&\bestcell{54.7}&\bestcell{53.1}&\bestcell{30.4}&\cellcolor{secondbest}27.6&\bestcell{57.5}&\bestcell{41.9}&\bestcell{43.6} \\
    \midrule
    & & \multicolumn{8}{c}{\textit{Proj-2d@5px}} & \\ \midrule
    \multirow{2}{*}{-}
    & PoseCNN & 34.6 & 15.1 & 10.4 & 31.8 & 7.4 & 1.9 & 13.8 & 23.1 & 17.2 \\ 
    & PVNet & 69.1 & 86.1 & 65.1 & 61.4 & 73.1 & 8.4 & 55.4 & 69.8 & 61.1\\
    \midrule
    \multirow{4}{*}{5}
    & OnePose++ & - & 0.0 & 0.0 & 0.0 & - & - & - & 0.0 & - \\
    & Gen6D &\cellcolor{thirdbest}13.4&\cellcolor{thirdbest}12.2&\cellcolor{thirdbest}0.5&\cellcolor{thirdbest}0.4&\cellcolor{thirdbest}9.6&\cellcolor{thirdbest}1.7&\cellcolor{thirdbest}5.7&\cellcolor{thirdbest}7.0&\cellcolor{thirdbest}6.3 \\
    & Gen6D$^{\dagger}$ &\cellcolor{secondbest}23.4&\cellcolor{secondbest}13.2&\cellcolor{secondbest}2.1&\cellcolor{secondbest}3.7&\cellcolor{secondbest}12.1&\bestcell{8.5}&\cellcolor{secondbest}14.1&\bestcell{15.6}&\cellcolor{secondbest}11.6 \\
    & \textbf{Ours} &\bestcell{41.7}&\bestcell{31.2}&\bestcell{10.4}&\bestcell{8.8}&\bestcell{42.8}&\cellcolor{secondbest}1.8&\bestcell{26.6}&\cellcolor{secondbest}14.5&\bestcell{21.9} \\
    \midrule
    \multirow{4}{*}{25}
    & OnePose++ & 0.0 & 0.0 & 8.9 & 0.0 & 6.3 &\cellcolor{secondbest}7.2& 0.0 & 2.5 & 3.1 \\
    & Gen6D &\cellcolor{thirdbest}37.3&\cellcolor{thirdbest}28.9&\cellcolor{thirdbest}18.6&\cellcolor{thirdbest}2.2&\cellcolor{thirdbest}36.5&\cellcolor{thirdbest}3.9&\cellcolor{thirdbest}11.4&\cellcolor{thirdbest}48.0&\cellcolor{thirdbest}23.4 \\
    & Gen6D$^{\dagger}$ &\cellcolor{secondbest}53.8&\cellcolor{secondbest}41.1&\cellcolor{secondbest}28.9&\cellcolor{secondbest}18.1&\cellcolor{secondbest}41.3&\bestcell{7.7}&\cellcolor{secondbest}32.3&\cellcolor{secondbest}65.2&\cellcolor{secondbest}36.1 \\
    & \textbf{Ours} &\bestcell{59.7}&\bestcell{58.9}&\bestcell{54.7}&\bestcell{27.8}&\bestcell{56.3}& 1.9 &\bestcell{53.6}&\bestcell{71.4}&\bestcell{47.9} \\
    \bottomrule
    \end{tabular}
    }
    \caption{
        \textbf{Comparison on the Occluded LINEMOD dataset.} ADD(s)-0.1d and Proj-2d@5px are reported. $^{\dagger}$Indicates that ground-truth detection results were provided for Gen6D. Objects in \textit{italic} were included in the Gen6D training set. We've highlighted the \colorbox{best}{\textbf{best}} and \colorbox{secondbest}{second-best} and \colorbox{thirdbest}{third-best} results.
    }
    \label{tab:LINEMOD-O}
\end{table}

For Occluded LINEMOD, as shown in Table~\ref{tab:LINEMOD-O}, our method consistently performs better than both OnePose++ and Gen6D under all conditions, which shows the limitations of the OnePose family in handling occlusions. Even when Gen6D is fine-tuned on a subset of LINEMOD and provided with ground-truth detection results, it still faces significant challenges in occlusion scenarios. Furthermore, our method attains performance comparable to or exceeding the instance-level method PVNet on some objects, demonstrating its robustness as an RGB-based generalizable approach in handling occlusions.

\subsection{Results on YCB-Video}
\label{sec:ycb}
The YCB-Video dataset is challenging for object pose estimation due to occlusions, motion blur, and variations in lighting. For the \textbf{Sparse Database}, Table~\ref{tab:YCB-V-Sparse} shows that our method significantly outperforms OnePose++ using only five reference images. With 16 reference images, our method achieves much higher performance, outperforming Gen6D with ground-truth detection 11.4\% both in ADD-S and ADD metrics.
\begin{table}[htbp]
    \centering
    \resizebox{\columnwidth}{!}{
    \begin{tabular}{l|cc|cc|cc|cc|cc}
    \toprule
    & \multicolumn{2}{c}{\textbf{Gen6D}$^{\ddagger}$} & \multicolumn{2}{c}{\textbf{Gen6D}$^{\dagger}$} & \multicolumn{2}{c}{\textbf{OnePose++}} & \multicolumn{2}{c}{\textbf{Ours}} & \multicolumn{2}{c}{\textbf{Ours}}  \\
    \cmidrule(lr){2-3} \cmidrule(lr){4-5} \cmidrule(lr){6-7} \cmidrule(lr){8-9} \cmidrule(lr){10-11}
    Ref. images  & \multicolumn{2}{c}{16} & \multicolumn{2}{c}{16} & \multicolumn{2}{c}{16} & \multicolumn{2}{c}{5} & \multicolumn{2}{c}{16} \\
    Metrics & ADD-S & ADD & ADD-S & ADD & ADD-S & ADD & ADD-S & ADD & ADD-S & ADD \\
    \midrule
    002\_master\_chef\_can & \cellcolor{secondbest}75.6 & \cellcolor{thirdbest}23.8 & 69.5 & \bestcell{29.4} & 46.1 & 17.4 & \bestcell{77.6} & \cellcolor{secondbest}28.9 & \cellcolor{thirdbest}71.4 & 22.3 \\
    003\_cracker\_box & 7.2 & 1.7 & \cellcolor{thirdbest}41.5 & \cellcolor{thirdbest}17.2 & 6.9 & 0.7 & \bestcell{72.3} & \bestcell{45.8} & \cellcolor{secondbest}69.1 & \cellcolor{secondbest}36.3 \\
    004\_sugar\_box & 10.4 & 4.6 & \bestcell{54.8} & \bestcell{26.2} & 1.0 & 0.1 & \cellcolor{thirdbest}47.6 & \cellcolor{thirdbest}17.2 & \cellcolor{secondbest}50.9 & \cellcolor{secondbest}21.9 \\
    005\_tomato\_soup\_can & \cellcolor{secondbest}63.3 & \cellcolor{secondbest}36.6 & 58.6 & \cellcolor{thirdbest}34.5 & - & - & \cellcolor{thirdbest}60.4 & 16.6 & \bestcell{81.5} & \bestcell{51.2} \\
    006\_mustard\_bottle & 39.9 & 19.4 & \cellcolor{secondbest}86.8 & \cellcolor{secondbest}57.2 & 46.7 & 30.6 & \cellcolor{thirdbest}84.1 & \cellcolor{thirdbest}50.1 & \bestcell{87.4} & \bestcell{76.4} \\
    007\_tuna\_fish\_can & \bestcell{87.7} & \bestcell{50.7} & \cellcolor{secondbest}85.2 & \cellcolor{thirdbest}47.8 & 0.0 & 0.0 & 36.9 & 13.7 & \cellcolor{thirdbest}77.4 & \cellcolor{secondbest}48.3 \\
    008\_pudding\_box & 19.1 & 1.9 & \cellcolor{thirdbest}50.9 & \cellcolor{thirdbest}28.2 & 0.0 & 0.0 & \bestcell{90.5} & \bestcell{80.8} & \cellcolor{secondbest}76.3 & \cellcolor{secondbest}66.0 \\
    009\_gelatin\_box & 40.8 & 19.3 & \cellcolor{secondbest}60.3 & \cellcolor{secondbest}38.7 & 0.5 & 0.2 & \cellcolor{thirdbest}54.7 & \cellcolor{thirdbest}34.0 & \bestcell{90.2} & \bestcell{81.9} \\
    010\_potted\_meat\_can & 54.7 & 31.8 & \cellcolor{thirdbest}58.5 & \cellcolor{thirdbest}40.2 & - & - & \cellcolor{secondbest}63.4 & \bestcell{52.7} & \bestcell{67.5} & \cellcolor{secondbest}49.6 \\
    011\_banana & 10.7 & 4.0 & \bestcell{37.2} & \cellcolor{secondbest}6.3 & 0.0 & 0.0 & \cellcolor{secondbest}32.4 & \cellcolor{thirdbest}5.1 & \cellcolor{thirdbest}30.5 & \bestcell{7.4} \\
    019\_pitcher\_base & 44.3 & 12.0 & \bestcell{80.3} & \cellcolor{secondbest}58.9 & 27.1 & 15.9 & \cellcolor{secondbest}80.2 & \bestcell{60.7} & \cellcolor{thirdbest}70.3 & \cellcolor{thirdbest}40.7 \\
    021\_bleach\_cleanser & 18.6 & 11.7 & \cellcolor{secondbest}61.6 & \cellcolor{thirdbest}39.6 & 42.0 & 32.2 & \cellcolor{thirdbest}58.1 & \cellcolor{secondbest}40.9 & \bestcell{84.9} & \bestcell{73.8} \\
    024\_bowl & 13.3 & \cellcolor{thirdbest}3.9 & \bestcell{45.9} & \bestcell{5.7} & 10.5 & 0.3 & \cellcolor{secondbest}40.0 & 3.2 & \cellcolor{thirdbest}35.2 & \cellcolor{secondbest}4.8 \\
    025\_mug & \cellcolor{thirdbest}73.7 & 29.3 & 72.7 & \cellcolor{thirdbest}41.1 & 3.7 & 0.9 & \bestcell{88.9} & \bestcell{76.4} & \cellcolor{secondbest}83.1 & \cellcolor{secondbest}65.8 \\
    035\_power\_drill & 5.6 & 0.9 & \cellcolor{thirdbest}39.7 & 9.2 & 22.4 & \cellcolor{thirdbest}12.8 & \cellcolor{secondbest}57.8 & \bestcell{42.2} & \bestcell{60.6} & \cellcolor{secondbest}38.3 \\
    036\_wood\_block & 10.9 & \cellcolor{secondbest}1.8 & \cellcolor{thirdbest}16.1 & \cellcolor{thirdbest}1.4 & 11.1 & 0.7 & \bestcell{33.6} & \bestcell{2.8} & \cellcolor{secondbest}20.5 & 0.3 \\
    037\_scissors & 1.3 & 0.1 & \bestcell{39.3} & \bestcell{17.8} & 0.0 & 0.0 & \cellcolor{thirdbest}16.9 & \cellcolor{secondbest}7.1 & \cellcolor{secondbest}17.5 & \cellcolor{thirdbest}4.1 \\
    040\_large\_marker & 30.8 & 20.7 & \cellcolor{thirdbest}49.4 & \cellcolor{thirdbest}39.0 & 0.0 & 0.0 & \bestcell{75.0} & \bestcell{61.9} & \cellcolor{secondbest}68.3 & \cellcolor{secondbest}56.4 \\
    051\_large\_clamp & 28.7 & 7.2 & \cellcolor{thirdbest}53.2 & \cellcolor{thirdbest}17.1 & 1.7 & 0.2 & \bestcell{68.7} & \bestcell{26.5} & \cellcolor{secondbest}66.8 & \cellcolor{secondbest}24.7 \\
    052\_extra\_large\_clamp & 6.5 & 2.1 & \cellcolor{thirdbest}36.8 & \cellcolor{secondbest}8.3 & 4.4 & 0.4 & \cellcolor{secondbest}49.7 & \cellcolor{thirdbest}5.9 & \bestcell{63.2} & \bestcell{22.4} \\
    061\_foam\_brick & 49.2 & \cellcolor{thirdbest}22.9 & \cellcolor{secondbest}60.4 & \bestcell{36.9} & 0.0 & 0.0 & \bestcell{65.3} & \cellcolor{secondbest}29.1 & \cellcolor{thirdbest}51.8 & 22.5 \\
    \midrule
    MEAN & 33.0 & 14.6 & \cellcolor{thirdbest}55.2 & \cellcolor{thirdbest}28.6 & 11.8 & 5.9 & \cellcolor{secondbest}59.8 & \cellcolor{secondbest}31.8 & \bestcell{66.6} & \bestcell{40.0} \\
    \bottomrule
    \end{tabular}
    }
    \caption{
        \textbf{Performance of the YCB-Video dataset on Sparse Database.} $^{\dagger}$Indicates that ground-truth detection results were provided for Gen6D. $^{\ddagger}$Indicates that the background was removed using ground-truth masks to improve Gen6D's detection results. We've highlighted the \colorbox{best}{\textbf{best}}, \colorbox{secondbest}{second-best}, and \colorbox{thirdbest}{third-best} results.
    }
    \label{tab:YCB-V-Sparse}
\end{table}

For the remaining two reference databases, average comparison results are provided in Table~\ref{tab:YCB-V-Avg}. To showcase the advantages of our approach better, we compare it with Gen6D and OnePose++, both of which use dense references (200 images sampled from the database); more detailed results are available in the Supplementary Material. While OnePose++ performs better with the dense database, Gen6D does not benefit from the dense samples due to its sensitivity to occlusion and low-quality reference views. In contrast, our method maintains stable performance with only five reference views, and the occlusion-minimizing reference database further enhances its accuracy.

\begin{table}[htbp]
    \centering
    \resizebox{\columnwidth}{!}{
    \begin{tabular}{c c c c c}
    \toprule
    \textbf{Ref. database} & \textbf{Ref. images} & \textbf{Method} & \textbf{ADD} & \textbf{ADD-S} \\
    \midrule
    \multirow{4}{*}{Most-overlapping}
    & 200 & OnePose++ & 24.5 & 43.3 \\
    & 200 & Gen6D & 14.6 & 29.1 \\
    & 200 & Gen6D$\dagger$ & 22.9 & 51.1 \\
    & 5 & \textbf{Ours} & \textbf{35.4} & \textbf{65.6} \\
    \midrule
    \multirow{4}{*}{Occlusion-minimizing}
    & 200 & OnePose++ & 22.6 & 41.7 \\
    & 200 & Gen6D & 12.2 & 26.7 \\
    & 200 & Gen6D$\dagger$ & 20.2 & 50.2 \\
    & 5 & \textbf{Ours} & \textbf{37.8} & \textbf{66.9} \\
    \bottomrule
    \end{tabular}
    }
    \caption{
        \textbf{Average Performance on YCB-Video dataset with different reference databases.} ADD and ADD-S are reported. Gen6D$^{\dagger}$ indicates that the method benefits from ground-truth detection results. The best results are shown in \textbf{bold}.
    }
    \label{tab:YCB-V-Avg}
\end{table}

\subsection{Results on OnePose-LowTexture}
We also conduct experiments on the OnePose-LowTexture dataset, which contains textureless objects (see Table~\ref{tab:OnePose-L-ADD}). Our method outperforms both OnePose++ and Gen6D using only 10 reference images (the minimum required for OnePose++ to reconstruct all objects) and remains competitive against the full reference results. Gen6D, not trained on this object-centric dataset, struggles with detection due to the similar scales between the reference and query views. This results in poor performance even when using ground-truth detections. More detailed results on OnePose and OnePose-LowTexture are provided in the Supplementary Material.

\label{sec:onepose}
\begin{table}[htbp]
    \centering
    \resizebox{\columnwidth}{!}{
    \begin{tabular}{l c cccccccc c}
    \toprule
    \textbf{Methods} & \textbf{Ref. images} & \textbf{0700} & \textbf{0706} & \textbf{0714} & \textbf{0721} & \textbf{0727} & \textbf{0732} & \textbf{0736} & \textbf{0740} & \textbf{Avg.} \\ 
    \midrule
    PVNet & - & 12.3 &\cellcolor{secondbest}90.0&\cellcolor{thirdbest}68.1& 67.6 &\cellcolor{thirdbest}95.6&\cellcolor{thirdbest}57.3& 61.3 &\cellcolor{thirdbest}49.3&\cellcolor{thirdbest}62.7 \\
    OnePose++ & Full &\bestcell{89.5}&\bestcell{99.1}&\bestcell{97.2}&\bestcell{92.6}&\bestcell{98.5}&\cellcolor{secondbest}79.5&\bestcell{97.2}&\cellcolor{secondbest}57.6&\bestcell{88.9} \\
    OnePose++ & 10 &\cellcolor{thirdbest}44.9& 64.9 & 56.5 &\cellcolor{thirdbest}78.8& 88.1 & 56.0 &\cellcolor{thirdbest}71.0& 3.7 & 58.0 \\
    Gen6D$^{\dagger}$ & 10 & 16.0 & 11.7 & 10.6 & 32.3 & 29.0 & 20.5 & 23.9 & 25.4 & 21.2 \\
    \textbf{Ours} & 10 &\cellcolor{secondbest}71.4&\cellcolor{thirdbest}85.6&\cellcolor{secondbest}84.4&\cellcolor{secondbest}88.8&\cellcolor{secondbest}96.2&\bestcell{92.6}&\cellcolor{secondbest}96.1&\bestcell{77.6}&\cellcolor{secondbest}86.6 \\
    \bottomrule
    \end{tabular}
    }
    \caption{\textbf{Performance comparison on OnePose-LowTexture dataset.} ADD-0.1d is reported. We've highlighted the \textbf{\colorbox{best}{best}}, \colorbox{secondbest}{second-best}, and \colorbox{thirdbest}{third-best} results.}
    \label{tab:OnePose-L-ADD}
\end{table}

\subsection{Analysis}
\label{sec:exp-analysis}
\begin{figure}[htbp]
    \centering
    \includegraphics[width=0.73\columnwidth]{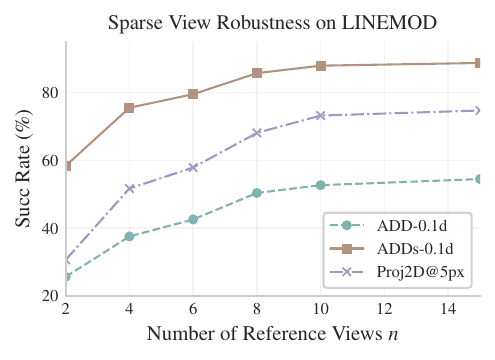}
    \caption{The trend of performance on the LINEMOD dataset as the number of reference views changes.}
    \label{fig:num_ref}
\end{figure}

\noindent \textbf{Effects of Number of Reference Images.} We evaluate the impact of varying the number of reference images on the LINEMOD dataset using the Proj2D@5px, ADD-0.1d, and ADDs-0.1d metrics. In this experiment, we directly feed different quantities into the network without any selection process to highlight the effect of the number of reference images. As shown in Fig.~\ref{fig:num_ref}, our method can estimate a coarse object pose even with only two reference images. With more reference images, the box estimation transformer effectively leverages the additional information to predict box corners more accurately.

\begin{figure}[htbp]
    \centering
    \includegraphics[width=0.62\columnwidth]{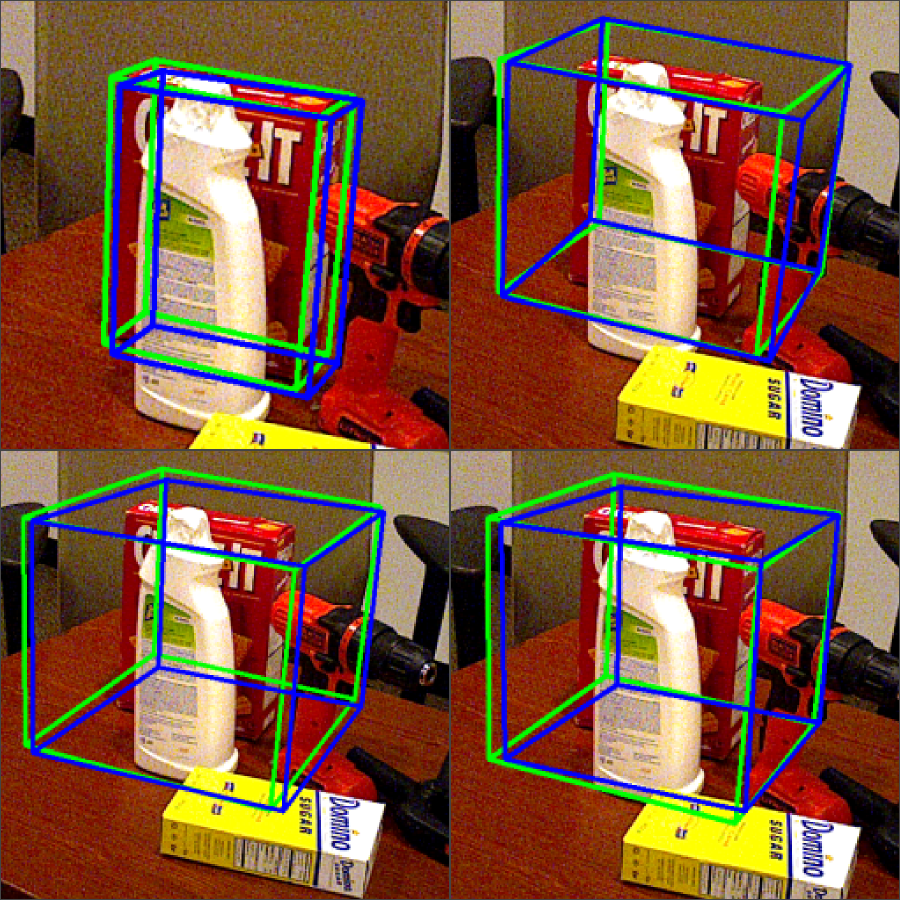}
    \caption{\textbf{Qualitative results on different bounding boxes.} Top left: ground-truth object bounding box; Top right and others: bounding boxes recovered from DUSt3R using five reference images from three different reference databases introduced in Sec.~\ref{sec:setup}.}
    \label{fig:box_robustness}
\end{figure}

\noindent \textbf{Effects of Noisy 3D Bounding Boxes.} We use DUSt3R to recover the object bounding boxes by filtering points with multiview detections. This process can introduce noise in the recovered boxes, so we evaluate our method's robustness to variations in bounding box quality. As shown in Fig.~\ref{fig:box_robustness}, our approach remains robust across different bounding boxes. Furthermore, experiments on the LINEMOD dataset with various bounding box sources (see Table~\ref{tab:box_robustness}) confirm the method's adaptation to noisy box estimates.

\begin{table}
    \centering
    \resizebox{\columnwidth}{!}{
    \begin{tabular}{c c c c}
    \toprule
    \textbf{Bounding Box Source} & \textbf{Ref. Images} & \textbf{ADD(s)-0.1d} & \textbf{Proj2D@5px} \\
    \midrule
    OnePose++ & 5 & 49.4 & \textbf{52.7} \\
    DUSt3R   & 5 & \textbf{51.3} & 43.8 \\
    GT       & 5 & 49.3 & 51.8 \\
    \bottomrule
    \end{tabular}
    }
    \caption{\textbf{Performance on LINEMOD using different bounding box sources.} The best results are shown in \textbf{bold}.}
    \label{tab:box_robustness}
\end{table}

\noindent \textbf{Effects of Real Data Training.} As shown in Table~\ref{tab:real-data-affects}, incorporating training on a simple object-centric dataset like OnePose can effectively enhance the performance of our approach on other challenging real-world datasets.

\begin{table}[htbp]
    \centering
    \resizebox{\columnwidth}{!}{
    \begin{tabular}{c c c c}
    \toprule
    \textbf{Real Data Training} & \textbf{Ref. Images} & \textbf{ADD(s)-0.1d} & \textbf{Proj2D@5px} \\
    \midrule
    w/o & 5 & 40.6 & 25.4 \\
    w   & 5 & \textbf{51.3} & \textbf{43.8} \\
    \bottomrule
    \end{tabular}
    }
    \caption{\textbf{Performance on LINEMOD with different training datasets.} The best results are shown in \textbf{bold}.}
    \label{tab:real-data-affects}
\end{table}

\noindent \textbf{Running time.} On an Intel i7-13700KF CPU and NVIDIA RTX 4090 GPU, our method processes a query image in approximately 17~ms using 5 reference images for pose inference. Specifically, it takes about 5.7~ms for the DINOv2 encoder, 11.1~ms for the bounding box decoder, and 0.15~ms for pose recovery, confirming its real-time performance. Prior to inference, reconstructing 5 reference images with DUSt3R (including model loading, reconstruction, and point cloud filtering) requires roughly 5.3~s. This process is conducted offline \emph{for once} and does not affect real-time inference.

%% file: sec/5_conclusion.tex
\section{Conclusion}
\label{sec:conclusion}

In this paper, we introduced BoxDreamer, a novel framework for generalizable object pose estimation that leverages object bounding box corners as an intermediate representation. Extensive experiments demonstrate that BoxDreamer significantly improves pose estimation performance, especially in challenging scenarios such as sparse view inputs and occlusions. Moreover, the efficient transformer decoder enables real-time performance, enhancing the method's practical applicability.

\noindent \textbf{Limitations and Future Work.} Although BoxDreamer shows promising results, it still struggles with estimating the poses of symmetric objects. While the framework can handle dense view inputs, this leads to higher memory usage. A key future direction is to leverage dense inputs for improved accuracy without additional time or memory costs. Moreover, since object detection and pose estimation are treated separately, integrating 2D object detection with 3D bounding box corner prediction would further enhance the method's practicality.

\newpage

\section*{Acknowledgment} 
This work was partially supported by the Major Program of Xiangjiang Laboratory (No. 24XJJCYJ01004), NSFC (No. 62402427, NO. U24B20154), Zhejiang Provincial Natural Science Foundation of China (No. LR25F020003), Ant Research, Zhejiang University Education Foundation Qizhen Scholar Foundation, and Information Technology Center and State Key Lab of CAD\&CG, Zhejiang University.

The authors would also like to take this opportunity to thank RoboSense (Suteng Innovation Technology Co., Ltd) for providing the AC1 (Active Camera) sensor, which was used to collect some of the data for our experiment and to conduct validations.